\documentclass[preprint,12pt,authoryear]{elsarticle}

\usepackage{amssymb}    
\usepackage{amsthm}     
\usepackage{amsmath}    
\usepackage{lineno}     
\usepackage{booktabs}   
\usepackage{rotating}   
\usepackage{subfigure}  
\usepackage[export]{adjustbox}  
\usepackage{booktabs}   
\usepackage{multirow}   
\usepackage[dvipsnames]{xcolor}         
\usepackage{threeparttable}             
\usepackage[margin=0.85in]{geometry}    
\usepackage[skip=0.6\baselineskip]{caption} 
\usepackage{array}      
\newcolumntype{L}[1]{>{\raggedright\let\newline\\\arraybackslash\hspace{0pt}}m{#1}}
\newcolumntype{C}[1]{>{\centering\let\newline\\\arraybackslash\hspace{0pt}}m{#1}}
\newtheorem{assumption}{Assumption}     
\usepackage{lipsum}                     
\usepackage{epsfig}                     
\usepackage[makeroom]{cancel}           

\usepackage{hyperref}                   
\hypersetup{colorlinks=true}

\journal{...}

\begin{document}

\begin{frontmatter}

\title{Spatio-temporally complementary feature propagation on graphs for longitudinal AADT estimation}

\author[a]{Linghang Sun\corref{cor1}} \ead{lisun@ethz.ch}
\author[b]{Qishen Zhou}
\author[a]{Michail A. Makridis}
\author[a]{Anastasios Kouvelas}

\affiliation[a]{organization={Institute for Transport Planning and Systems, ETH Zurich},
            city={Zurich},
            postcode={8093}, 
            country={Switzerland}}

\affiliation[b]{organization={School of Transportation, Jilin University},
            city={Changchun},
            postcode={130012}, 
            country={China}}

\cortext[cor1]{Corresponding author.}

\begin{abstract}
The estimation of Annual Average Daily Traffic (AADT) is vital for transportation planning and infrastructure maintenance, yet obtaining accurate values for an entire urban network across multiple years remains challenging due to the high cost and spatial sparsity of physical sensors. This research proposes a novel spatio-temporally complementary feature propagation framework that leverages the strengths of two distinct data sources: spatially sparse but temporally dense loop detector data, and a spatially complete but temporally sparse macroscopic transportation model. The methodology highlights a feature propagation algorithm on directed graphs, formulated as a Poisson energy minimization considering residues. The standard binary adjacency matrix is replaced with flow ratio matrices to capture real-world vehicle turn ratios at intersections. Validated in the city of Zurich, the algorithm demonstrates high computational efficiency, achieving convergence within minutes. Results indicate that the framework effectively reconciles theoretical models with empirical ground truths, yielding a normalized mean absolute error below $10\%$. This scalable approach provides a feasible solution for spatio-temporal network-wide AADT estimation through combining real-world limited sensor coverage and traffic models.

\end{abstract}


\begin{keyword}
Poisson energy minimization \sep multimodal feature propagation \sep longitudinal AADT estimation
\end{keyword}

\end{frontmatter}


\section{Introduction}
\label{sec:introduction}

\begin{figure}[hbt!]
  \includegraphics[width=0.9\textwidth]{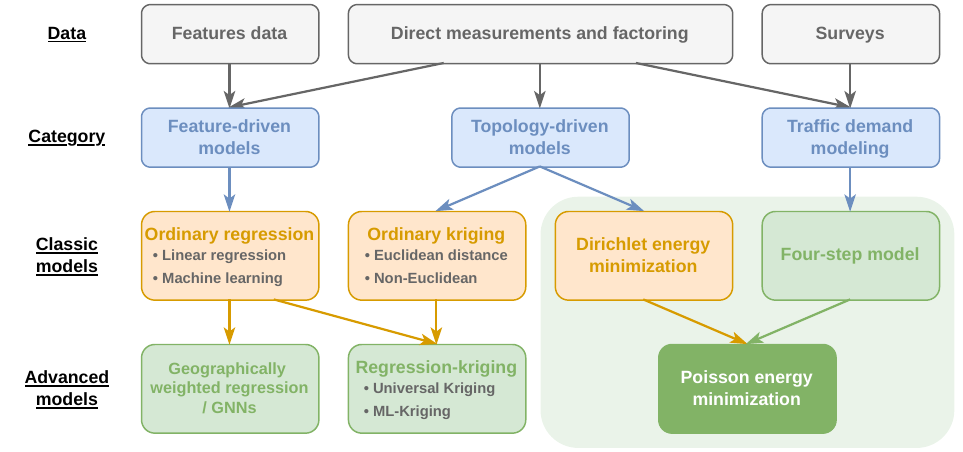}
  \centering
  \caption{Established AADT estimation algorithms.}
  \label{fig:aadt_methods}
\end{figure}

Understanding traffic distribution across a network is essential for effective urban infrastructure and transportation planning. For applications ranging from pavement maintenance and environmental assessments to intelligent transportation systems and safety analysis \citep{huynh2020development, archondo-callao_maintenance_2011, chen_modeling_2016}, the fundamental metric is Annual Average Daily Traffic (AADT), defined as the volume of traffic passing through a road on an average day over an investigated year. While AADT can be directly recorded using sensors like inductive loops, radar, and video cameras, data coverage in modern cities is still sparse and highly uneven. As sensor equipment and maintenance are expensive, dense monitoring is typically restricted to major arterial roads and signalized intersections, leaving most of the urban network unobserved. To overcome this lack of direct measurement, researchers have designed various algorithms to estimate AADT for unmonitored road segments, as illustrated in Fig \ref{fig:aadt_methods}. Models in orange are the classic approaches that provide methodological foundations that may fail hard when applied directly without adaptation to AADT estimation. Models in light green are the state-of-the-art algorithms that achieve satisfactory results for estimating static AADT. The proposed longitudinal AADT estimation algorithm, shown in dark green, leverages a combination of Poisson energy minimization and transport modeling for propagating AADT information from observable locations to unobserved locations.

\textbf{Direct measurements and factoring} serve as the foundation for most of the established algorithms. While measurements largely come from permanent sensors, unobserved locations can also be monitored through portable counters for a certain number of days or weeks \citep{gastaldi_estimation_2014}. Such measurements will then be factored based on recommendations provided by local authorities, such as \cite{fhwa2016tmg} in the US. The rising penetration rate of floating car in recent years may also provide a good estimation of AADT when the penetration can be clearly defined. 

The observed measurements of AADT statistics will be fed into an established estimation algorithm, which generally falls into one of the three categories: feature-driven models, topology-driven models, and transportation models:

\textbf{Feature-driven models} refer to the regression-based methods, such as ordinary regression and lately machine learning algorithms \citep{yangNewEfficientRegression2014, geccheleAdvancesUncertaintyTreatment2012, zhangStatewideTruckVolume2023}. On top of direct AADT measurements, additional feature information about these observable road sections should also be available, including but not limited to road type, number of lanes, land use, speed limit, and population density. By computing the correlation between AADT and the feature variables at the observable roads, the AADT of unobserved locations can be interpolated. Geographically weighted regression (GWR) is an advanced spatially-aware feature-driven AADT estimation algorithm. GWR confines the correlation with a Gaussian kernel and is often found to yield superior results \citep{zhaoUsingGeographicallyWeighted2004, pulugurthaModelingAADTLocal2021}. Another breakthrough in this feature-driven branch is the application of graph neural networks (GNNs) \citep{zhenAnalyzingImportanceNetwork2024}. However, both GWR and GNN are data-hungry to produce meaningful results.

\textbf{Topology-driven models} leverage the topology structure of the network or simply the geospatial relationship of the roads to achieve AADT estimation, and kriging is a major family of such geostatistical algorithms \citep{krige1951statistical}, such as ordinary kriging and universal kriging. Ordinary Kriging predicts values at unmeasured locations by calculating a weighted average of known nearby data points based on Euclidean or topological distance and clustering, where a variogram is applied to minimize the estimation error. However, when implementing ordinary Kriging, minor roads tend to get heavily influenced by the surrounding major roads and have abnormally high AADT estimates \citep{selbySpatialPredictionTraffic2013}. To overcome the issue, regression-kriging takes related features into account and has been widely applied in both AADT estimation and AADT prediction \citep{songTrafficVolumePrediction2019}. Recent advances in learning on graphs provide another potential novel methodology for AADT estimation \citep{rossi_unreasonable_2022, zhou_network-wide_2025}. Through Dirichlet energy minimization (DEM), information can be propagated from observable locations to those unobservable. However, similar to ordinary kriging, if directly applied to AADT estimation, DEM often leads to high estimates on the neighborhood roads as well. 

\textbf{Traffic demand modeling} estimates AADT from a different angle by capturing real-world mobility patterns. Following a traditional or modified four-step model, the origin, destination, route choice, and mode choice of all the trips on an average day are modeled \citep{aprontiFourstepTravelDemand2018a}, which naturally yields AADT statistics on the road network. The simulated model may also be calibrated iteratively with real-world measurements to achieve high fidelity \citep{horni2016multi}. To achieve an accurate representation of such mobility patterns, demand modeling requires massive survey data, socioeconomic data, and land use information, among others \citep{zhenAnalyzingImportanceNetwork2024}. Therefore, traffic demand models can be heavily resource-intensive, and building one new model per year to update AADT can be monetarily prohibitive.

It is important to point out that with the methodological advancement, the boundary between feature-driven models and topology-driven models has blurred, and modern algorithms take the advantages of both ends. Such observation inspired the development of the proposed longitudinal AADT estimation algorithm based on DEM. Likewise, we incorporate DEM and traffic demand modeling together and reformulate the AADT estimation as a Poisson energy minimization problem. Although the proposed algorithm has the limitation of requiring a high-fidelity demand model in the first place, the advantages over the established algorithms are also essential and multifold:

\begin{itemize}[-]
    \item The algorithm produces longitudinal AADT data, which is of great necessity for safety research.
    \item Both AADT and intersection turn ratios can be estimated to study potential rat-running behavior.
    \item The algorithm echoes the real-world physical constraint of the traffic flow conservation law.
    \item The algorithm is very sample efficient and requires no road feature data.

\end{itemize}

\begin{figure}[hbt!]
  \includegraphics[width=1\textwidth]{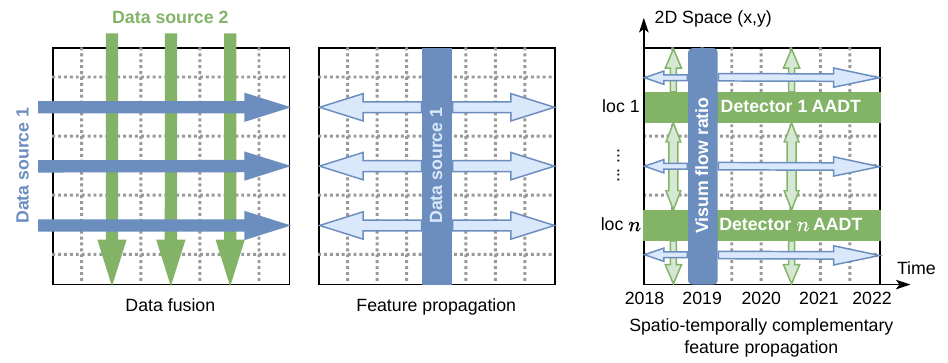}
  \centering
  \caption{Conceptualization of the proposed algorithm.}
  \label{fig:fusion}
\end{figure}

Despite the exceptionally abundant amount of data in the digital era, different types of sensors and detectors are designed to observe physical phenomena from diverse perspectives. We then choose loop detector measurements and demand modeling, because one of them represents the trend over space and the other over time. The relation between various data sources can be further exploited to harness valuable information that is blinded to a single information source. The process of combining data sources from different data acquisition frameworks is commonly known as multimodal data fusion \citep{lahat_multimodal_2015}, and feature propagation indicates techniques to interpolate or extrapolate unobservables through observed measurements. We further build on these concepts and propose the concept of spatio-temporally complementary feature propagation in this work, as illustrated in Fig. \ref{fig:fusion}, demonstrating deducing feature values on both the temporal and spatial dimensions. While this research mainly focuses on AADT estimation, the flow ratio can also be easily estimated across the years.

\section{Methodology}
\label{sec:methodology}

Unlike highway networks, which are a closed environment and widely considered as following the traffic flow conservation law, urban road networks, on the other hand, generally do not follow such physical constraints. In this work, we force conservability by computing the inbound and outbound residues on each road section and taking flow ratios at each intersection as the underlying rule for the diffusion. We first develop the feature propagation algorithm on a generalized directed graph following a Poisson energy functional scheme in Section \ref{sec:unconservability}, considering the residues, where the standard binary adjacency matrix is adopted. We demonstrate that by calculating the gradient flow, the feature set converges to the energy minimum. Then the framework is extended to the transportation context in Section \ref{sec:flow} by redefining the adjacency matrix in terms of flow ratios, which is computed as the proportion of total traffic moving toward each available direction at an intersection. By incorporating residue vectors and flow ratio matrix, we conclude the formulation of the feature propagation algorithm following the traffic flow conservation law.

\subsection{Feature propagation following Poisson energy} 
\label{sec:unconservability}

\begin{figure}[hbt!]
  \includegraphics[width=1\textwidth]{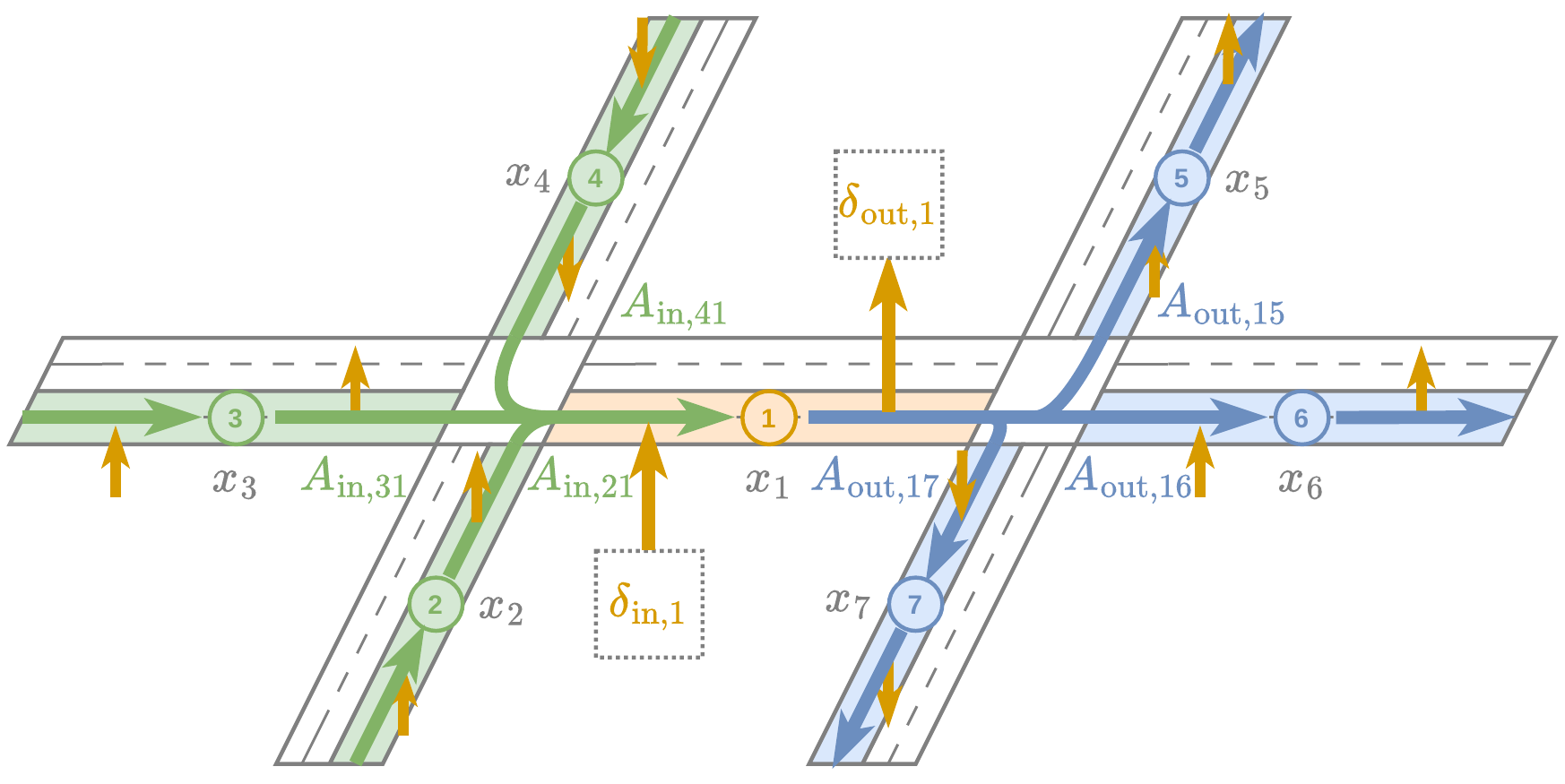}
  \centering
  \caption{An illustration of a link-based graph and feature propagation.}
  \label{fig:graph}
\end{figure}

Leveraging a dual graph representation, we abstract the urban road network into a directed graph $G = (V, E)$, where each node $v \in V$ represents a directed road link, such that $V=\{v_1, v_2, \dots, v_n\}$. The edge set $E = \{e_1, e_2, \dots, e_m\}$ represents the permissible flow at intersections, indicating an edge $e_k = (v_i, v_j)$ exists if traffic can legally maneuver from road link $v_i$ into link $v_j$. Such directed connectivity $A_{ij}$ can be defined with an $n \times n$ adjacency matrix $\mathbf{A}$, whereas $\mathbf{D}^{\mathrm{in}} = \mathrm{diag}(\sum_i A_{i1}, \dots, \sum_i A_{in})$ and $\mathbf{D}^{\mathrm{out}} = \mathrm{diag}(\sum_j A_{1j}, \dots, \sum_j A_{nj})$ each denotes the in-degree and the out-degree matrix and are diagonal matrix of the adjacency matrix's column-sum and row-sum, respectively. To model the additional source and sink at the upstream and downstream on each node, as shown in Fig. \ref{fig:graph}, we further introduce $\boldsymbol{\delta}^{\mathrm{in}}$ and $\boldsymbol{\delta}^{\mathrm{out}}$ as inbound and outbound residue vectors. In the context of urban road networks, the residual vector can represent private drives, parking garages with multiple entrances and exits, and links at the network border. Likewise, when considering $\mathbf{A}$ as the outflow adjacency matrix on the directed graph, its transpose $\mathbf{A}^\top$ serves as the inflow adjacency matrix, effectively representing the same connectivity from the perspective of the target nodes. Hence, the graph Laplacian can be defined as $\mathbf{L} = (\mathbf{D}^{\mathrm{in}} + \mathbf{D}^{\mathrm{out}}) - (\mathbf{A} + \mathbf{A}^\top)$, which can be decomposed into the inbound graph Laplacian $\mathbf{L}^{\mathrm{in}} = \mathbf{D}^{\mathrm{in}} - \mathbf{A}^\top$ and the outbound Laplacian $\mathbf{L}^{\mathrm{out}} = \mathbf{D}^{\mathrm{out}} - \mathbf{A}$ with $\mathbf{L} = \mathbf{L}^{\mathrm{in}} + \mathbf{L}^{\mathrm{out}}$.

The Poisson energy functional on the graph $G$ is defined as $E_P(\mathbf{x}, G) = E_D(\mathbf{x}, G) - \ell(\mathbf{x}, G)$, which consists of the Dirichlet energy $E_D(\mathbf{x}, G) = \frac{1}{2}\mathbf{x}^{\top} \mathbf{L} \mathbf{x}$ and an additional linear functional $\ell(\mathbf{x}, G) = \mathbf{x}^{\top} \boldsymbol{\delta} $ with the residual vector defined following an inbound-and-outbound formulation $\boldsymbol{\delta} = \mathbf{L}^{\mathrm{in}}\boldsymbol{\delta}^{\mathrm{in}} + \mathbf{L}^{\mathrm{out}}\boldsymbol{\delta}^{\mathrm{out}}$. While the Dirichlet energy term acts as a regularizer to minimize the energy variation between each adjacent node, the linear functional constrains the regularization through external residuals.

Let $V_o \subseteq V$ denote the subset of nodes equipped with sensors. The corresponding node feature vector $\mathbf{x} \in \mathbb{R}^n$ can then be partitioned into two complementary subsets $\mathbf{x}_o$ and $\mathbf{x}_u$, representing the observable and unobservable features, respectively. The features for the unobserved nodes $\mathbf{x}_u$ are subsequently determined by minimizing $E_P$ subject to the observed features $\mathbf{x}_o$. By partitioning the graph into observed and unobserved node sets, the feature vector, adjacency matrix, graph laplacian, and residual vector can be expressed in block form:

\begin{equation}
\mathbf{x} =
\begin{bmatrix}
{\mathbf{x}}_o \\
{\mathbf{x}}_u
\end{bmatrix},
\; \mathbf{A} =
\begin{bmatrix}
\mathbf{A}_{oo} & \mathbf{A}_{ou} \\
\mathbf{A}_{uo} & \mathbf{A}_{uu}
\end{bmatrix},
\; \mathbf{D}^{\mathrm{in}} =
\begin{bmatrix}
{\mathbf{D}^{\mathrm{in}}_{o}} \\
{\mathbf{D}^{\mathrm{in}}_{u}}
\end{bmatrix},
\; \mathbf{D}^{\mathrm{out}} =
\begin{bmatrix}
{\mathbf{D}^{\mathrm{out}}_{o}} \\
{\mathbf{D}^{\mathrm{out}}_{u}}
\end{bmatrix},
\; \mathbf{L} =
\begin{bmatrix}
\mathbf{L}_{oo} & \mathbf{L}_{ou} \\
\mathbf{L}_{uo} & \mathbf{L}_{uu}
\end{bmatrix}
\; \boldsymbol{\delta} =
\begin{bmatrix}
\boldsymbol{\delta}_o \\
\boldsymbol{\delta}_u
\end{bmatrix}
\end{equation}

The theoretical foundations of Dirichlet energy $E_D$ minimization for feature propagation are explored in \cite{rossi_unreasonable_2022} on undirected graphs and in \cite{zhou_network-wide_2025} on directed graphs. Building upon these foundations, this work develops a feature propagation algorithm based on the Poisson energy functional $E_P$ on the directed graph.

The feature set converges to the energy minimum through the gradient flow of the Poisson functional with specific boundary and initial conditions:
\begin{subequations}
\begin{align}
\nonumber
\dot{\mathbf{x}}(t) & = -\nabla E_P(\mathbf{x}(t), G)
\\
\label{eq:gradient}
& = - \mathbf{L} \mathbf{x}(t) + \boldsymbol{\delta}
\\
\text{(BC)} \quad 
\mathbf{x}_o(t) &= \mathbf{x}_o, \quad \forall t \ge 0
\\
\text{(IC)} 
\quad \mathbf{x}(0) &= 
\begin{bmatrix}
\mathbf{x}_o  \\
\mathbf{x}_u(0) 
\end{bmatrix}
\end{align}
\end{subequations}

The rate of gradient vanishes as the system reaches equilibrium with $\mathbf{x}^* = \lim_{t \to \infty} \mathbf{x}(t)$, implying that the feature vector is governed by the Poisson equation $\mathbf{L} \mathbf{x}^* = \boldsymbol{\delta}$. This Poisson energy optimization problem can be approximated through Jacobi iteration and resetting known features at each iteration. When discretizing Eq. \ref{eq:gradient}, such diffusion can be achieved through numerical iterations $k=1,2,\dots$, with preconditioner $\mathbf{H}$ as the node-dependent step size:

\begin{align}
\label{eq:discretize}
\mathbf{x}^{(k+1)} - \mathbf{x}^{(k)} = 
- \mathbf{H} \mathbf{L} \mathbf{x}^{(k)} + \mathbf{H} \boldsymbol{\delta}
\end{align}

When the preconditioner is set to $\mathbf{H} = (\mathbf{D}^{\mathrm{in}} + \mathbf{D}^{\mathrm{out}})^{-1}$, and since by definition $\mathbf{L} = (\mathbf{D}^{\mathrm{in}} + \mathbf{D}^{\mathrm{out}}) - (\mathbf{A} + \mathbf{A}^\top)$, Eq. \ref{eq:discretize} can be further rewritten as: 

\begin{align}
\label{eq:diffusion}
\mathbf{x}^{(k+1)} = 
(\mathbf{D}^{\mathrm{in}} + \mathbf{D}^{\mathrm{out}})^{-1} (\mathbf{A} + \mathbf{A}^\top) \mathbf{x}^{(k)} + (\mathbf{D}^{\mathrm{in}} + \mathbf{D}^{\mathrm{out}})^{-1} \boldsymbol{\delta}
\end{align}

We incorporate the boundary and initial conditions into the gradient flow in Eq. \ref{eq:gradient}:

\begin{equation}
\begin{bmatrix}
\dot{\mathbf{x}}_o(t) \\
\dot{\mathbf{x}}_u(t)
\end{bmatrix}
=
-
\begin{bmatrix}
0 & 0 \\
\mathbf{L}_{uo} & \mathbf{L}_{uu}
\end{bmatrix}
\begin{bmatrix}
\mathbf{x}_o \\
\mathbf{x}_u(t)
\end{bmatrix}
+
\begin{bmatrix}
0 \\
\boldsymbol{\delta}_u
\end{bmatrix}
\end{equation}

When expressed in discretized form with the preconditioner $\mathbf{H}$:

\begin{equation}
\label{eq:matrix}
\mathbf{x}^{(k+1)}
= 
 \begin{bmatrix}
\mathbf{I} & 0 \\
- \mathbf{H}\mathbf{L}_{uo} & \mathbf{I} - \mathbf{H}\mathbf{L}_{uu}
\end{bmatrix}
\mathbf{x}^{(k)}
+
\begin{bmatrix}
0 \\
\mathbf{H} \boldsymbol{\delta}_u
\end{bmatrix}
\end{equation}

\subsection{Feature propagation considering flow ratio} 
\label{sec:flow}

In transportation networks, instead of being simply binary, the adjacency matrix can be designed to carry more real-world physical meanings, such as leveraging the Euclidean distance \citep{rahmani_graph_2023}. Let the adjacency matrix $\mathbf{A}$ be redefined as the outbound flow ratio matrix $\mathbf{A}^{\mathrm{out}}$, describing the percentage of flow towards each admissible direction. Note that the inbound flow ratio $\mathbf{A}^{\mathrm{in}}$ is defined differently compared to $\mathbf{A}^{\top}$. We introduce a flow matrix $\mathbf{F}$ with $F_{ij}$ representing the traffic flow from node $i$ to node $j$, so that $\mathbf{F} = \mathrm{diag}(\mathbf{x} + \boldsymbol{\delta}^{\mathrm{out}}) \mathbf{A}^{\mathrm{out}} = (\mathrm{diag}(\mathbf{x} + \boldsymbol{\delta}^{\mathrm{in}}) \mathbf{A}^{\mathrm{in}})^{\top}$ when the feature vector $\mathbf{x}$ is known. Hence, the inbound flow ratio is:

\begin{align}
\label{eq:upstream}
\mathbf{A}^{\mathrm{in}} = 
\mathrm{diag}(\mathbf{x} + \boldsymbol{\delta}^{\mathrm{in}})^{-1} (\mathrm{diag}(\mathbf{x} + \boldsymbol{\delta}^{\mathrm{out}}) \mathbf{A}^{\mathrm{out}})^{\top}
\end{align}

It should be noted that by such conversion, the new graph Laplacian $\mathbf{L}_{\mathrm{flow}}$ is no longer symmetric in the Euclidean space. As the directed flow-ratio matrices generally satisfy $\mathbf{A}^{\mathrm{in}} \neq (\mathbf{A}^{\mathrm{out}})^\top$, the update rule cannot be strictly interpreted as the classical gradient descent of a quadratic Poisson energy. However, as demonstrated by the flow matrix $\mathbf{F}$, the underlying physical exchanges are perfectly balanced when scaled by the local traffic volumes. Consequently, the algorithm effectively operates in a state-dependent metric space, allowing us to characterize this process as a non-Euclidean Poisson-inspired energy minimization. 

Since $\mathbf{A}^{\mathrm{out}}$ is defined as the outbound flow ratio matrix and the sum of each row becomes 1, rendering the out-degree matrix the same as an identity matrix $\mathbf{D}^{\mathrm{out}} = \mathbf{I}$. Unlike Section \ref{sec:unconservability}, where the inbound degree matrix is derived from the columns of the adjacency matrix, here $\mathbf{A}^{\mathrm{in}}$ is defined as an independent row-stochastic matrix. Therefore, its row-sum degree matrix likewise becomes an identity matrix, $\mathbf{D}^{\mathrm{in}} = \mathbf{I}$. Note for boundary links where flow enters or exits the observed network, this flow conservation is perfectly maintained by the previously introduced inbound and outbound residue vectors, which explicitly account for links at the network border. Therefore, Eq. \ref{eq:diffusion} can be rewritten as:

\begin{align}
\label{eq:flow}
\mathbf{x}^{(k+1)} = \frac{1}{2} (\mathbf{A}^{\mathrm{in}} + \mathbf{A}^{\mathrm{out}}) \mathbf{x}^{(k)} + \frac{1}{2} (\mathbf{I} - \mathbf{A}^{\mathrm{in}}) \boldsymbol{\delta}^{\mathrm{in}} + \frac{1}{2} (\mathbf{I} - \mathbf{A}^{\mathrm{out}}) \boldsymbol{\delta}^{\mathrm{out}} 
\end{align}

Because $\mathbf{A}^{\mathrm{in}}$ and $\mathbf{A}^{\mathrm{out}}$ are stochastic matrices with each row summing to $1$, their combination forms a non-expansive operator $\frac{1}{2} (\mathbf{A}^{\mathrm{in}} + \mathbf{A}^{\mathrm{out}})$. This guarantees that the non-Euclidean propagation remains unconditionally stable, converging monotonically to a steady-state fixed point without oscillations when validated on a single source dataset. A similar proof can be found in \cite{zhou_network-wide_2025}.

\subsection{Incorporation of the data sources and assumptions} 
\label{sec:assumptions}

Two data sources are combined to achieve the spatio-temporally complementary data fusion for the AADT estimation in the city of Zurich. One data source is loop detector data, which can be considered the ground truth of AADT. However, despite being \textbf{temporally complete}, loop detectors are \textbf{spatially sparse} due to the installation and maintenance costs. The other data source is the integrated transportation model (Gesamtverkehrsmodell in German, GVM in short) for the canton of Zurich in the format of PTV Visum \citep{zh_gvm_website}. The GVM model is a multimodal, macroscopic transport demand model that integrates travel behavior, network supply, and land-use data to simulate traffic flows across regions. It is developed using extensive data sources, including travel surveys, census data, and traffic counts. Unlike loop detector data, the GVM model is \textbf{spatially complete} by showing the AADT of every single road in the investigated region. The model is \textbf{temporally sparse} due to the high costs of conducting comprehensive surveys to develop the model, with the latest model developed in 2019. Other data sources can also be adopted to calculate the flow ratio matrix, such as floating car data from TomTom or other commercial service providers. However, the costs for retrieving such data over multiple years can be intimidating.

The GVM model is utilized to get the residue vectors $\boldsymbol{\delta}^{\mathrm{in}}$ and $\boldsymbol{\delta}^{\mathrm{out}}$, which are an integral part of the model, and to calculate the flow ratio matrices $\mathbf{A}^{\mathrm{in}}$ and $\mathbf{A}^{\mathrm{out}}$. The detectors provide the ground truth of AADT for the observable links $\mathbf{x}_o$. We further denote $\mathbf{A}^{\mathrm{GVM}} = (\mathbf{A}^{\mathrm{in}} + \mathbf{A}^{\mathrm{out}})/2$ and $\boldsymbol{\delta}^{\mathrm{GVM}} = ( (\mathbf{I} - \mathbf{A}^{\mathrm{in}}) \boldsymbol{\delta}^{\mathrm{in}} + (\mathbf{I} - \mathbf{A}^{\mathrm{out}}) \boldsymbol{\delta}^{\mathrm{out}}) /2$, and the original diffusion model Eq. \ref{eq:matrix} becomes:

\begin{equation}
\label{eq:fusion}
\begin{bmatrix}
\mathbf{x}^{\mathrm{det}} \\
\mathbf{x}_u^{(k+1)}
\end{bmatrix}
= 
 \begin{bmatrix}
\mathbf{I} & 0 \\
\mathbf{A}^{\mathrm{GVM}}_{uo} & \mathbf{A}^{\mathrm{GVM}}_{uu}
\end{bmatrix}
\begin{bmatrix}
\mathbf{x}^{\mathrm{det}} \\
\mathbf{x}_u^{(k)}
\end{bmatrix}
+
\begin{bmatrix}
0 \\
\boldsymbol{\delta}^{\mathrm{GVM}}_{}
\end{bmatrix}
\end{equation}

The spatio-temporally complementary data fusion between the detector data and the GVM model is built on the following two assumptions, which will be validated in Section \ref{sec:results}.

\begin{assumption}
\label{ass:accurate}
GVM provides a high-fidelity approximation of empirical mobility patterns.
\end{assumption}

While GVM is developed under extensive surveys, well-established methodologies, and elaborated calibration, the model does not perfectly reflect the absolute ground truth of traffic. Nevertheless, it can be considered as a rather accurate representation of the real-world mobility. 

\begin{assumption} 
\label{ass:constant}
The flow ratio maintains mostly temporal consistency over multi-year horizons.
\end{assumption}

Although both AADT on the nodes and flow ratio on the edges can be retrieved or deduced from GVM, AADT can shift drastically across the year, especially considering the years under the influence of COVID-19 \citep{marraImpactCOVID19Pandemic2022}. On the contrary, the flow ratio is heavily dependent on population density, points of interest, and hence the evolution is much slower and gradual. It should be noted that this assumption does not contradict the investigation of the accident exposure shift and the rat-running behavior. The essence of the algorithm is not one source of data correcting the other, but rather calibrating each other to minimize the Poisson energy. As the flow ratio is only fixed in the year of the GVM model, the minor change of such values due to the implementation of the countermeasures in the following years will be flagged and calibrated by detector data. 

\section{Results}
\label{sec:results}

\begin{figure}[hbt!]
  \includegraphics[width=0.75\textwidth]{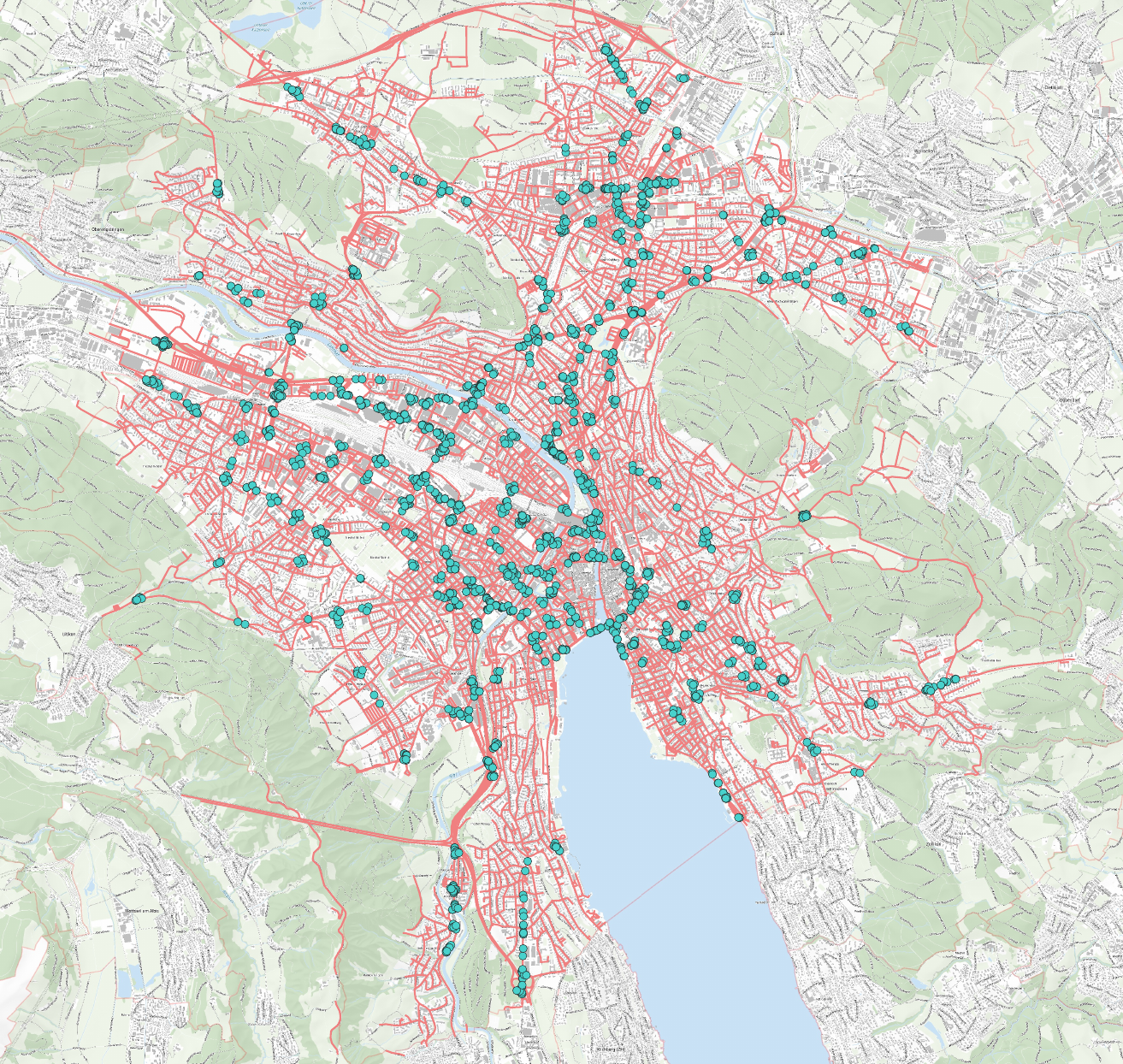}
  \centering
  \caption{Visualization of the GVM and the loop detectors.}
  \label{fig:map}
\end{figure}

The GVM model and detector measurements are in distinct data formats, and the quality of results depends significantly on data preprocessing and matching, which is achieved in QGIS in this work with an illustrative example as shown in Fig. \ref{fig:map}. The GVM model is generated based on the year of $2019$ in the city of Zurich for 45100 directed edges covering all transportation modes, including pedestrian roads, waterways, railways, and so on. We cleansed the irrelevant modes and kept 9947 road segments. On the other hand, the detector data is available for 5 years between 2018 and 2022. There are a total of 1943 detectors installed within the city. As AADT measures the traffic volume at the link level instead of at the lane level, these detectors are checked one by one manually to form the correct detector group so that the correct AADT can be computed for those observable locations. We apply the OSMnx Python package to simplify the network by removing the nodes that are not truly intersections \cite{boeingModelingAnalyzingUrban2025}. The detector coverage is computed to be $16.0\%$ in the Zurich network.

\begin{figure}[hbt!]
  \includegraphics[width=0.6\textwidth]{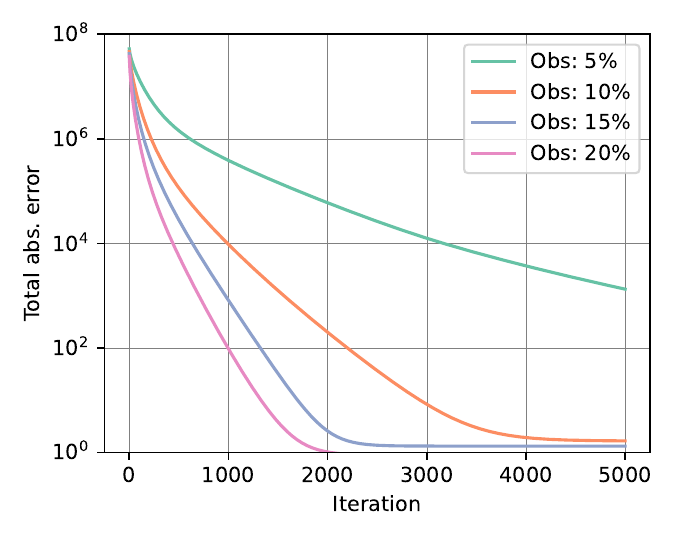}
  \centering
  \caption{GVM self-validation with the proposed algorithm.}
  \label{fig:self}
\end{figure}

First, to demonstrate the effectiveness of the proposed algorithm in Eq. \ref{eq:fusion}, we start with a self-validation study with the GVM model alone by replacing $\mathbf{x}^{\mathrm{det}}$ with $\mathbf{x}^{\mathrm{GVM}}$, in which a varying percentage of features is masked. Fig. \ref{fig:self} illustrates how the total absolute error summed from the unobserved nodes changes through diffusion iterations. Results showcased that with a small percentage of observable nodes, the feature gradually propagates into the whole graph with decreasing absolute error. Note that the y-axis is in $\mathrm{log}$ scale, indicating that a linear relationship on the plot represents an exponential decay in the error value. When given a sufficient number of iterations, the error converges to an extremely small value. The algorithm is computationally efficient and can achieve the $5000$ iterations within minutes.

\begin{figure}[hbt]
  \centering
    \subfigure[AADT ratio in $\mathrm{log}_2(\mathrm{GVM/det})$ scale.]{%
      {\includegraphics[width=0.48\textwidth]{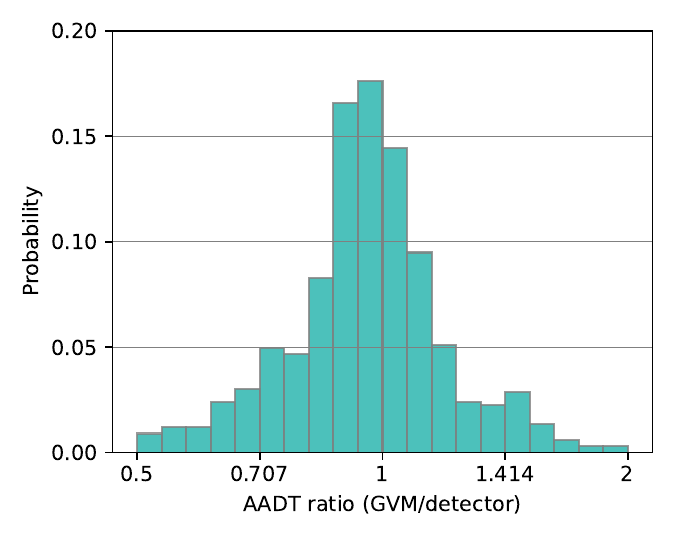}}\label{fig:ratio}}
    \subfigure[Coefficient of variation over $2018-2022$]{%
      {\includegraphics[width=0.48\textwidth]{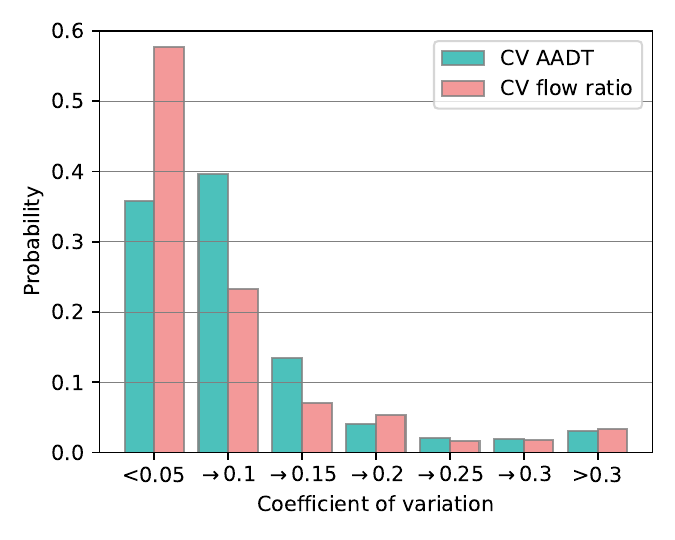}}\label{fig:cv}}
    \caption{Validations of assumptions.} 
\end{figure}

Following the successful demonstration of the algorithm with self-validation, we apply the spatio-temporally complementary data fusion algorithm to the combined GVM model and loop detector data for the city of Zurich. This process begins with validating the two primary assumptions introduced in Section \ref{sec:assumptions}. For the matched nodes with observable features, Fig. \ref{fig:ratio} represents the AADT ratio between GVM and detectors. Note we apply a $\mathrm{log}_2(x)$ scale instead of a simple ratio to avoid the imbalance between positive and negative ratios. Therefore, the left end of the axis indicates that GVM is half of detector AADT, whereas the right end shows that detector AADT is half of GVM. With $2^{0.2} \approx 1.15$, approximately $60\%$ of all the observable node features have a difference within $15\%$, validating Assumption \ref{ass:accurate}.

To showcase that flow ratio is an indicator more consistent in comparison to AADT throughout the years, we compute the Coefficient of Variation (CV) of both AADT and flow ratio based on the ground truth detector readings throughout the $5$ years, and demonstrate the distribution of CV in Fig. \ref{fig:cv}. It is clear to observe that most observable flow ratio has a CV below $0.05$ while the mode of CV for AADT is $0.05-0.10$. Such a difference in CV validates Assumption \ref{ass:constant}.

\begin{figure}[hbt]
  \centering
    \subfigure[Diffusion curves and cross validation.]{%
      {\includegraphics[width=0.48\textwidth]{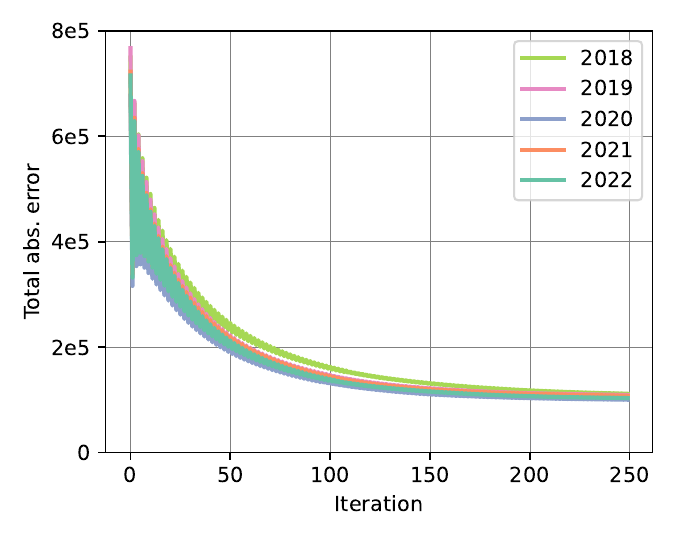}}\label{fig:cross_error}}
    \subfigure[NMAE across the year $2018-2022$.]{%
      {\includegraphics[width=0.48\textwidth]{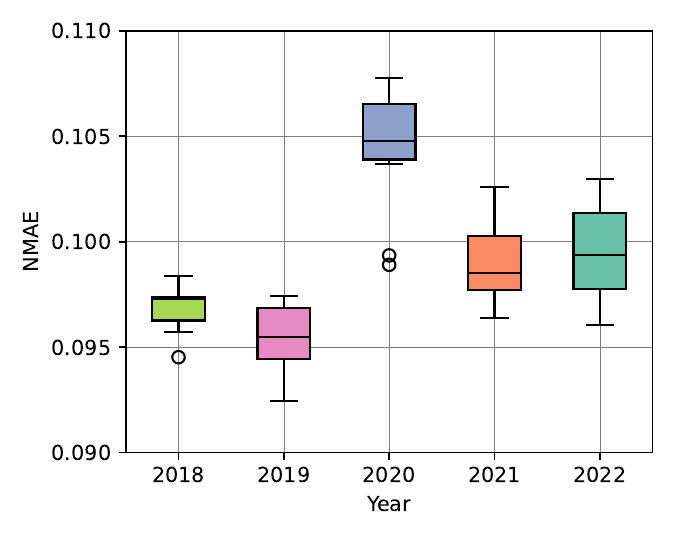}}\label{fig:boxplot}}
    \caption{Data fusion combining GVM and detector data over $2018-2022$.} 
\end{figure}

Fig. \ref{fig:cross_error} illustrates the results of spatio-temporally complementary data fusion combining both the GVM model from $2019$ and the detector data from year $2018$ to $2022$. We split all observable features into two subsets, with $80\%$ leveraged for feature propagation and the other $20\%$ for validation. Since the validation subset is selected randomly, a total of $10$ diffusion runs are implemented for each year. The illustrated results are an average of the $10$ diffusion runs. It should be noted that the feature propagation is set to be 1000 iterations, with only the first 250 iterations displayed in Fig. \ref{fig:cross_error} for better visualization of the diffusion process before convergence. In comparison to Fig. \ref{fig:self}, the total absolute error converges to a significant level, due to the intrinsic distinction between the GVM model and real-world detector data. Such variation also leads to the strong feature oscillation at the beginning of the diffusion process, and gradually attenuates over iterations. The diffusion curves across the years show similar characteristics and largely overlap with each other. Fig. \ref{fig:boxplot} further details the error of each year. Each box represents the normalized mean absolute error (NMAE) at the end of $1000$ iterations, averaged from the $10$ diffusion runs in the investigated year. Overall, the NMAE lies within $0.10$, demonstrating satisfactory estimation of AADT. Another interesting phenomenon is the shifting of NMAE. Since the GVM model is generated based on the survey and data in $2019$, the NMAE in $2019$ is expected to be the lowest. Although in Assumption \ref{ass:constant}, we assume the flow ratio to be rather consistent over the years, the error still grows from $2019$ to $2021$ and $2022$. Year $2020$ is an exceptional case with an NMAE of $0.105$ potentially due to the influence of COVID-19, showing that the pandemic not only drastically reduced the mobility willingness of residents but also choices of destination and perception of utility.

\section{Conclusion}
\label{sec:conclusion}
In this work, we propose a two-dimensional feature propagation algorithm for data fusion on graphs with spatio-temporally complementary data sources for cross-year AADT estimation. By minimizing the Poisson energy on a directed graph, we solve the feature propagation problem considering residues on the nodes. A key contribution of this research is the adaptation of the algorithm to the transportation domain by replacing the standard binary adjacency matrix with a flow ratio matrix, representing the vehicle turn ratio toward admissible directions at each intersection. The data sources leveraged in this research are loop detector data and the GVM transportation model developed by the canton of Zurich. First, the effectiveness of the proposed algorithm is confirmed based on the self-validation of the GVM model. Then, the underlying assumptions that GVM is a relatively accurate representation of real-world mobility, as well as the flow ratio being more consistent over the years, compared to AADT. Finally, we implement the spatio-temporally complementary data feature propagation with both detector data and the GVM model. Results demonstrate that the error converges to a low value, with NMAE showing an average error lower than $10\%$. The algorithm is highly scalable due to its computational efficiency. 

Several limitations of this research need to be addressed. As the algorithm developed in this work relies on the quality of data and model, researchers and practitioners should exercise their own discretion over the quality. A transportation demand model is resource-intensive and thus not commonly accessible in every city. However, the spatio-temporally complementary feature propagation algorithm can be model-agnostic, meaning any data source providing flow ratio information can be leveraged, such as TomTom, HERE, or other floating car data providers.

The proposed algorithm has several strengths that would also heavily benefit the planned future works. This research serves as a methodological tool for an ongoing study to examine the impact of Tempo 30 speed regulations on traffic accident rates in Zurich. Since detector coverage is relatively low in urban networks, the algorithm proposed in this work provides the high-resolution longitudinal AADT statistics as an accident exposure index, which is highly critical for such safety analyses. Although the validation demonstrates an NMAE under $10\%$, which can be considered as highly accurate for such task, another indispensable work to be carried out instantly is to demonstrate the performance of the proposed algorithm against established algorithms, such as Universal kriging and GWR. Nevertheless, neither universal kriging nor GWR can estimate both AADT and flow ratio at the same time, which is another core advantage of the spatio-temporally complementary feature propagation algorithm. The potential shift of the flow ratio may unveil whether speed limit reduction regulations effectively decrease the traffic accident rate or simply relocate the accidents to surrounding neighborhood roads.

\section*{Acknowledgements}
This research was funded by the Swiss State Secretariat for Education, Research, and Innovation (SERI) with contract number 25.00120, as part of the European Union's Horizon Europe project AntifragiCity with project ID 101203052. The authors would also like to thank the Amt für Mobilität of the canton of Zurich for providing the GVM Visum model, as well as Transcality and the Dienstabteilung Verkehr of the city of Zurich for providing the loop detector data.

\bibliographystyle{elsarticle-harv} 
\bibliography{references}

\begin{thebibliography}{23}
\expandafter\ifx\csname natexlab\endcsname\relax\def\natexlab#1{#1}\fi
\providecommand{\url}[1]{\texttt{#1}}
\providecommand{\href}[2]{#2}
\providecommand{\path}[1]{#1}
\providecommand{\DOIprefix}{doi:}
\providecommand{\ArXivprefix}{arXiv:}
\providecommand{\URLprefix}{URL: }
\providecommand{\Pubmedprefix}{pmid:}
\providecommand{\doi}[1]{\href{http://dx.doi.org/#1}{\path{#1}}}
\providecommand{\Pubmed}[1]{\href{pmid:#1}{\path{#1}}}
\providecommand{\bibinfo}[2]{#2}
\ifx\xfnm\relax \def\xfnm[#1]{\unskip,\space#1}\fi
\bibitem[{Apronti and Ksaibati(2018)}]{aprontiFourstepTravelDemand2018a}
\bibinfo{author}{Apronti, D.T.}, \bibinfo{author}{Ksaibati, K.}, \bibinfo{year}{2018}.
\newblock \bibinfo{title}{Four-step travel demand model implementation for estimating traffic volumes on rural low-volume roads in {{Wyoming}}}.
\newblock \bibinfo{journal}{Transportation Planning and Technology} \bibinfo{volume}{41}, \bibinfo{pages}{557--571}.
\newblock \DOIprefix\doi{10.1080/03081060.2018.1469288}.
\bibitem[{Archondo-Callao(2011)}]{archondo-callao_maintenance_2011}
\bibinfo{author}{Archondo-Callao, R.}, \bibinfo{year}{2011}.
\newblock \bibinfo{title}{Maintenance {Impact} on {Economic} {Evaluation} of {Upgrading} {Unsealed} {Roads}}.
\newblock \bibinfo{journal}{Transportation Research Record} \bibinfo{volume}{2203}, \bibinfo{pages}{151--159}.
\newblock \DOIprefix\doi{10.3141/2203-19}.
\bibitem[{Boeing(2025)}]{boeingModelingAnalyzingUrban2025}
\bibinfo{author}{Boeing, G.}, \bibinfo{year}{2025}.
\newblock \bibinfo{title}{Modeling and {{Analyzing Urban Networks}} and {{Amenities With OSMnx}}}.
\newblock \bibinfo{journal}{Geographical Analysis} \bibinfo{volume}{57}, \bibinfo{pages}{567--577}.
\newblock \DOIprefix\doi{10.1111/gean.70009}.
\bibitem[{Chen and Xie(2016)}]{chen_modeling_2016}
\bibinfo{author}{Chen, C.}, \bibinfo{author}{Xie, Y.}, \bibinfo{year}{2016}.
\newblock \bibinfo{title}{Modeling the effects of {AADT} on predicting multiple-vehicle crashes at urban and suburban signalized intersections}.
\newblock \bibinfo{journal}{Accident Analysis \& Prevention} \bibinfo{volume}{91}, \bibinfo{pages}{72--83}.
\newblock \DOIprefix\doi{10.1016/j.aap.2016.02.016}.
\bibitem[{{Federal Highway Administration}(2016)}]{fhwa2016tmg}
\bibinfo{author}{{Federal Highway Administration}}, \bibinfo{year}{2016}.
\newblock \bibinfo{title}{Traffic Monitoring Guide}.
\newblock \bibinfo{type}{Report} \bibinfo{number}{FHWA-PL-17-003}. United States Department of Transportation.
\bibitem[{Gastaldi et~al.(2014)Gastaldi, Gecchele and Rossi}]{gastaldi_estimation_2014}
\bibinfo{author}{Gastaldi, M.}, \bibinfo{author}{Gecchele, G.}, \bibinfo{author}{Rossi, R.}, \bibinfo{year}{2014}.
\newblock \bibinfo{title}{Estimation of {Annual} {Average} {Daily} {Traffic} from one-week traffic counts. {A} combined {ANN}-{Fuzzy} approach}.
\newblock \bibinfo{journal}{Transportation Research Part C: Emerging Technologies} \bibinfo{volume}{47}, \bibinfo{pages}{86--99}.
\newblock \DOIprefix\doi{10.1016/j.trc.2014.06.002}.
\bibitem[{Gecchele et~al.(2012)Gecchele, Rossi, Gastaldi and Kikuchi}]{geccheleAdvancesUncertaintyTreatment2012}
\bibinfo{author}{Gecchele, G.}, \bibinfo{author}{Rossi, R.}, \bibinfo{author}{Gastaldi, M.}, \bibinfo{author}{Kikuchi, S.}, \bibinfo{year}{2012}.
\newblock \bibinfo{title}{Advances in {{Uncertainty Treatment}} in {{FHWA Procedure}} for {{Estimating Annual Average Daily Traffic Volume}}}.
\newblock \bibinfo{journal}{Transportation Research Record} \bibinfo{volume}{2308}, \bibinfo{pages}{148--156}.
\newblock \DOIprefix\doi{10.3141/2308-16}.
\bibitem[{Horni et~al.(2016)Horni, Nagel and Axhausen}]{horni2016multi}
\bibinfo{author}{Horni, A.}, \bibinfo{author}{Nagel, K.}, \bibinfo{author}{Axhausen, K.W.}, \bibinfo{year}{2016}.
\newblock \bibinfo{title}{The multi-agent transport simulation MATSim}.
\newblock \bibinfo{publisher}{Ubiquity Press}.
\bibitem[{Huynh et~al.(2020)Huynh, Mullen, Gassman, Ziehl and Ahmed}]{huynh2020development}
\bibinfo{author}{Huynh, N.}, \bibinfo{author}{Mullen, R.}, \bibinfo{author}{Gassman, S.}, \bibinfo{author}{Ziehl, P.}, \bibinfo{author}{Ahmed, F.}, \bibinfo{year}{2020}.
\newblock \bibinfo{title}{Development of pavement design and investigation strategies for non-interstate routes} .
\bibitem[{{Kanton Zürich}(2019)}]{zh_gvm_website}
\bibinfo{author}{{Kanton Zürich}}, \bibinfo{year}{2019}.
\newblock \bibinfo{title}{Gesamtverkehrsmodell (gvm-zh)}.
\bibitem[{Krige(1951)}]{krige1951statistical}
\bibinfo{author}{Krige, D.G.}, \bibinfo{year}{1951}.
\newblock \bibinfo{title}{A statistical approach to some basic mine valuation problems on the witwatersrand}.
\newblock \bibinfo{journal}{Journal of the Southern African Institute of Mining and Metallurgy} \bibinfo{volume}{52}, \bibinfo{pages}{119--139}.
\bibitem[{Lahat et~al.(2015)Lahat, Adali and Jutten}]{lahat_multimodal_2015}
\bibinfo{author}{Lahat, D.}, \bibinfo{author}{Adali, T.}, \bibinfo{author}{Jutten, C.}, \bibinfo{year}{2015}.
\newblock \bibinfo{title}{Multimodal {Data} {Fusion}: {An} {Overview} of {Methods}, {Challenges}, and {Prospects}}.
\newblock \bibinfo{journal}{Proceedings of the IEEE} \bibinfo{volume}{103}, \bibinfo{pages}{1449--1477}.
\newblock \DOIprefix\doi{10.1109/JPROC.2015.2460697}.
\bibitem[{Marra et~al.(2022)Marra, Sun and Corman}]{marraImpactCOVID19Pandemic2022}
\bibinfo{author}{Marra, A.D.}, \bibinfo{author}{Sun, L.}, \bibinfo{author}{Corman, F.}, \bibinfo{year}{2022}.
\newblock \bibinfo{title}{The impact of {{COVID-19}} pandemic on public transport usage and route choice: {{Evidences}} from a long-term tracking study in urban area}.
\newblock \bibinfo{journal}{Transport Policy} \bibinfo{volume}{116}, \bibinfo{pages}{258--268}.
\newblock \DOIprefix\doi{10.1016/j.tranpol.2021.12.009}.
\bibitem[{Pulugurtha and Mathew(2021)}]{pulugurthaModelingAADTLocal2021}
\bibinfo{author}{Pulugurtha, S.S.}, \bibinfo{author}{Mathew, S.}, \bibinfo{year}{2021}.
\newblock \bibinfo{title}{Modeling {{AADT}} on local functionally classified roads using land use, road density, and nearest nonlocal road data}.
\newblock \bibinfo{journal}{Journal of Transport Geography} \bibinfo{volume}{93}, \bibinfo{pages}{103071}.
\newblock \DOIprefix\doi{10.1016/j.jtrangeo.2021.103071}.
\bibitem[{Rahmani et~al.(2023)Rahmani, Baghbani, Bouguila and Patterson}]{rahmani_graph_2023}
\bibinfo{author}{Rahmani, S.}, \bibinfo{author}{Baghbani, A.}, \bibinfo{author}{Bouguila, N.}, \bibinfo{author}{Patterson, Z.}, \bibinfo{year}{2023}.
\newblock \bibinfo{title}{Graph {Neural} {Networks} for {Intelligent} {Transportation} {Systems}: {A} {Survey}}.
\newblock \bibinfo{journal}{IEEE Transactions on Intelligent Transportation Systems} \bibinfo{volume}{24}, \bibinfo{pages}{8846--8885}.
\newblock \DOIprefix\doi{10.1109/TITS.2023.3257759}.
\bibitem[{Rossi et~al.(2022)Rossi, Kenlay, Gorinova, Chamberlain, Dong and Bronstein}]{rossi_unreasonable_2022}
\bibinfo{author}{Rossi, E.}, \bibinfo{author}{Kenlay, H.}, \bibinfo{author}{Gorinova, M.I.}, \bibinfo{author}{Chamberlain, B.P.}, \bibinfo{author}{Dong, X.}, \bibinfo{author}{Bronstein, M.M.}, \bibinfo{year}{2022}.
\newblock \bibinfo{title}{On the {Unreasonable} {Effectiveness} of {Feature} {Propagation} in {Learning} on {Graphs} {With} {Missing} {Node} {Features}}, in: \bibinfo{booktitle}{Proceedings of the {First} {Learning} on {Graphs} {Conference}}, \bibinfo{publisher}{PMLR}. pp. \bibinfo{pages}{11:1--11:16}.
\bibitem[{Selby and Kockelman(2013)}]{selbySpatialPredictionTraffic2013}
\bibinfo{author}{Selby, B.}, \bibinfo{author}{Kockelman, K.M.}, \bibinfo{year}{2013}.
\newblock \bibinfo{title}{Spatial prediction of traffic levels in unmeasured locations: Applications of universal kriging and geographically weighted regression}.
\newblock \bibinfo{journal}{Journal of Transport Geography} \bibinfo{volume}{29}, \bibinfo{pages}{24--32}.
\newblock \DOIprefix\doi{10.1016/j.jtrangeo.2012.12.009}.
\bibitem[{Song et~al.(2019)Song, Wang, Wright, Thatcher, Wu and Felix}]{songTrafficVolumePrediction2019}
\bibinfo{author}{Song, Y.}, \bibinfo{author}{Wang, X.}, \bibinfo{author}{Wright, G.}, \bibinfo{author}{Thatcher, D.}, \bibinfo{author}{Wu, P.}, \bibinfo{author}{Felix, P.}, \bibinfo{year}{2019}.
\newblock \bibinfo{title}{Traffic {{Volume Prediction With Segment-Based Regression Kriging}} and its {{Implementation}} in {{Assessing}} the {{Impact}} of {{Heavy Vehicles}}}.
\newblock \bibinfo{journal}{IEEE Transactions on Intelligent Transportation Systems} \bibinfo{volume}{20}, \bibinfo{pages}{232--243}.
\newblock \DOIprefix\doi{10.1109/TITS.2018.2805817}.
\bibitem[{Yang et~al.(2014)Yang, Wang and Bao}]{yangNewEfficientRegression2014}
\bibinfo{author}{Yang, B.}, \bibinfo{author}{Wang, S.G.}, \bibinfo{author}{Bao, Y.}, \bibinfo{year}{2014}.
\newblock \bibinfo{title}{New {{Efficient Regression Method}} for {{Local AADT Estimation}} via {{SCAD Variable Selection}}}.
\newblock \bibinfo{journal}{IEEE Transactions on Intelligent Transportation Systems} \bibinfo{volume}{15}, \bibinfo{pages}{2726--2731}.
\newblock \DOIprefix\doi{10.1109/TITS.2014.2318039}.
\bibitem[{Zhang and Chen(2023)}]{zhangStatewideTruckVolume2023}
\bibinfo{author}{Zhang, X.}, \bibinfo{author}{Chen, M.}, \bibinfo{year}{2023}.
\newblock \bibinfo{title}{Statewide {{Truck Volume Estimation Using Probe Vehicle Data}} and {{Machine Learning}}}.
\newblock \bibinfo{journal}{Transportation Research Record} \bibinfo{volume}{2677}, \bibinfo{pages}{588--601}.
\newblock \DOIprefix\doi{10.1177/03611981231157400}.
\bibitem[{Zhao and Park(2004)}]{zhaoUsingGeographicallyWeighted2004}
\bibinfo{author}{Zhao, F.}, \bibinfo{author}{Park, N.}, \bibinfo{year}{2004}.
\newblock \bibinfo{title}{Using {{Geographically Weighted Regression Models}} to {{Estimate Annual Average Daily Traffic}}}.
\newblock \bibinfo{journal}{Transportation Research Record} \bibinfo{volume}{1879}, \bibinfo{pages}{99--107}.
\newblock \DOIprefix\doi{10.3141/1879-12}.
\bibitem[{Zhen and Yang(2024)}]{zhenAnalyzingImportanceNetwork2024}
\bibinfo{author}{Zhen, H.}, \bibinfo{author}{Yang, J.J.}, \bibinfo{year}{2024}.
\newblock \bibinfo{title}{Analyzing the importance of network topology in {{AADT}} estimation: Insights from travel demand models using graph neural networks}.
\newblock \bibinfo{journal}{Transportation} \DOIprefix\doi{10.1007/s11116-024-10536-y}.
\bibitem[{Zhou et~al.(2025)Zhou, Zhang, Makridis, Kouvelas, Wang and Hu}]{zhou_network-wide_2025}
\bibinfo{author}{Zhou, Q.}, \bibinfo{author}{Zhang, Y.}, \bibinfo{author}{Makridis, M.A.}, \bibinfo{author}{Kouvelas, A.}, \bibinfo{author}{Wang, Y.}, \bibinfo{author}{Hu, S.}, \bibinfo{year}{2025}.
\newblock \bibinfo{title}{Network-{Wide} {Freeway} {Traffic} {Estimation} {Using} {Sparse} {Sensor} {Data}: {A} {Dirichlet} {Graph} {Auto}-{Encoder} {Approach}}.
\newblock \bibinfo{journal}{IEEE Transactions on Intelligent Transportation Systems} \bibinfo{volume}{26}, \bibinfo{pages}{22161--22177}.
\newblock \DOIprefix\doi{10.1109/TITS.2025.3610911}.

\end{thebibliography}
\end{document}